\documentclass[letterpaper, 10 pt, journal, twoside]{IEEEtran}
\usepackage[T1]{fontenc}
\usepackage[latin9]{inputenc}
\usepackage[dvipsnames]{xcolor}
\usepackage{array}
\usepackage{color}
\usepackage{float}
\usepackage{mathrsfs}
\usepackage{mathtools}
\usepackage{multirow}
\usepackage{textcomp}
\usepackage{amsmath}
\usepackage{amsthm}
\usepackage{amssymb}
\usepackage{stackrel}
\usepackage{graphicx}
\usepackage{wasysym}
\usepackage{verbatim}

\makeatletter

\floatstyle{ruled}
\newfloat{algorithm}{tbp}{loa}
\providecommand{\algorithmname}{Algorithm}
\floatname{algorithm}{\protect\algorithmname}

\theoremstyle{definition}
\newtheorem{example}{\protect\examplename}
\theoremstyle{definition}
\newtheorem{problem}{\protect\problemname}
\theoremstyle{definition}
\newtheorem{defn}{\protect\definitionname}
\theoremstyle{plain}
\newtheorem{thm}{\protect\theoremname}

\usepackage{cite}
\usepackage{algpseudocode}
\usepackage{setspace}
\usepackage{color}
\usepackage{tikz}
\tikzset{>=latex}
\IEEEoverridecommandlockouts

\makeatother

\usepackage{babel}
\providecommand{\definitionname}{Definition}
\providecommand{\examplename}{Example}
\providecommand{\problemname}{Problem}
\providecommand{\theoremname}{Theorem}

\usepackage{lipsum}

\def\BibTeX{{\rm B\kern-.05em{\sc i\kern-.025em b}\kern-.08em
		T\kern-.1667em\lower.7ex\hbox{E}\kern-.125emX}}
\begin{document}
\title{Exploiting Transformer in Sparse Reward Reinforcement Learning for Interpretable
Temporal Logic Motion Planning
\thanks{Manuscript received: March 3, 2023; Revised: May 8, 2023; Accepted: June 22, 2023. This paper was recommended for publication by
Editor Jens Kober upon evaluation of the Associate Editor and Reviewers'
comments. This work was supported in part by the National Natural Science Foundation of China under Grant 62173314 and U2013601. (Corresponding Author: Zhen Kan)}
\thanks{H. Zhang, H. Wang, and Z. Kan are with the
Department of Automation at the University of Science and Technology
of China, Hefei, Anhui, China, 230026.}
\thanks{Digital Object
Identifier (DOI): see top of this page.}
}
\author{Hao Zhang, Hao Wang, and Zhen Kan}

\markboth{IEEE Robotics and Automation Letters. Preprint Version. Accepted June, 2023}
{Zhang \MakeLowercase{\textit{et al.}}: Exploiting Transformer in Sparse Reward Reinforcement Learning}

\maketitle
\begin{abstract}
Automaton based approaches have enabled robots to perform various
complex tasks. However, most existing automaton based algorithms highly rely on the manually customized representation of states for the considered task, limiting its applicability in deep reinforcement learning algorithms. To address this issue,
by incorporating Transformer into reinforcement learning, we develop
a Double-Transformer-guided Temporal Logic framework (T2TL) that exploits
the structural feature of Transformer twice, i.e., first encoding
the LTL instruction via the Transformer module for efficient understanding
of task instructions during the training and then encoding the context
variable via the Transformer again for improved task performance.
Particularly, the LTL instruction is specified by co-safe LTL. As
a semantics-preserving rewriting operation, LTL progression is exploited
to decompose the complex task into learnable sub-goals, which
not only converts non-Markovian reward decision processes to Markovian
ones, but also improves the sampling efficiency by simultaneous learning
of multiple sub-tasks. An environment-agnostic LTL pre-training scheme
is further incorporated to facilitate the learning of the Transformer
module resulting in an improved representation of LTL. The
simulation results demonstrate the effectiveness of
the T2TL framework.

\global\long\def\prog{\operatorname{prog}}%
\global\long\def\argmax{\operatorname{argmax}}%
\global\long\def\argmin{\operatorname{argmin}}%
\end{abstract}

\section{Introduction}

\IEEEPARstart{O}{ne} of the ultimate goals in robotic learning is to let the robot
infer the key to the task completion. To enable such human-level intelligence,
the capability of comprehending the semantics of instructions and
evolving continuously via interactions with the environment is crucial.
Among numerous learning algorithms, reinforcement learning (RL) is
a sequential decision-making process that models dynamics of the interaction
as a Markov decision process (MDP) and focuses on learning the optimal
policy through exploration and exploitation \cite{Sutton2018}. Although RL based methods have enabled the robot to accomplish tasks from simple to complex
ones, an important yet challenging topic
is how the robot can enhance their understanding of instructions to
improve task completion. In particular, there are three main challenges:
1) unlike existing works with explicit task instructions and motion
constraints, how can the robot comprehend the nature of instructions
by its own to improve the task completion? 2) Since many practical
tasks require the robot to perform a series of logically organized
sub-tasks (e.g., cleaning rooms, organizing books and washing clothes
while avoiding collisions), resulting in a non-Markovian reward decision
process (NMRDP), how can the NMRDP be properly handled? 3) When solving
the complex task in a sparse reward environment, how can the robot
facilitate learning by leveraging the potential of
its representation module? 

Transformer was originally presented in \cite{vaswani2017attention}
for natural language processing and recently achieves remarkable
success in many fields. In \cite{dosovitskiy2020image}, a Vision
Transformer (ViT) framework is developed, which proposes patch embedding
for image preprocessing and performs better than state-of-the-art
CNNs. The work of \cite{chen2021decision} presents an effective combination
of RL and Transformer, which casts the traditional RL problem as a conditional
sequence modeling by leveraging the causally masked Transformer. The
structured features of Transformer are further incorporated in \cite{liu2022robot}
to improve robotic manipulation by capturing the spatio-temporal
relationship between the dual-arm movements. Despite recent progress, most of the existing methods
with Transformer mainly focus on natural language processing or computer
vision, lacking the guidance to drive robots towards task completion.
It is unclear how conventional Transformer can be combined with
RL to guide the agent to understand complex motion planning tasks
that consist of a series of sub-goals that need to be completed logically.

Due to the rich expressivity and capability, linear temporal logic
(LTL) is capable of describing a wide range of complex tasks composed
of logically organized sub-tasks \cite{Baier2008}.
By converting the LTL specification into an automaton, learning algorithms
are often exploited to facilitate the motion planning of robotic systems.
For instance, modular deep reinforcement learning is incorporated
with a limit deterministic generalized B\"uchi automaton (LDGBA) to
enable continuous motion planning of an autonomous dynamical system  \cite{Cai2021b}. Learning-based probabilistic motion planning
subject to the deterministic Rabin automaton (DRA) guideline in the
presence of environment and motion uncertainties is investigated in
\cite{Cai2021c}. Truncated LTL is leveraged to facilitate the reward
design in \cite{Li2019}, which can be converted into a finite-state
predicate automaton (FSPA) to improve the performance of reinforcement
learning in robotic planning. Similar to the automaton, reward machine
(RM) is proposed to offer dense rewards feedback in \cite{Icarte2022},
which can be translated from a variety of temporal logic specifications
to improve the sample efficiency of reinforcement learning methods.
However, most of these aforementioned methods highly rely on the representation
of system states in the form of either automaton or RM, which not
only grows exponentially with respect to the task complexity, but also are not effective for deep learning (i.e., the customized automaton states with sorted index or manually one-hot encoding in RM generally cannot facilitate the gradient propagation of neural networks).
When considering representing the LTL as a neural network, the work
of \cite{Kuo2020} exploits a compositional recurrent neural network
(RNN) as an encoder to train the learning agent to understand LTL
semantics. However, RNN generally suffers from high computational cost
due to its inherently sequential nature precluding parallelization.
In \cite{vaezipoor2021ltl2action}, the compositional syntax and the semantics
of LTL are exploited by the relational graph convolutional network
(R-GCN) to enable the generalization to new tasks. However, it cannot
offer interpretable guidance to the agent due to irregularity
of the R-GCN architecture \cite{yuan2022explainability}. 

To bridge the gap, we consider using Transformer
to encode the LTL specifications to provide a more appropriate
representation to improve the performance, and offer reasonable
interpretability for task completion.

The main contributions of this work are summarized
as follows:

1. To our best knowledge, this is the first work
that encodes LTL instructions by Transformer to accomplish a complex
task with Reinforcement Learning in a sparse reward environment,
whose representation not only yields better policy performance than traditional one-hot encoding or sorted index representations,
but also further provides reasonable interpretability for the agent's
motion planning.

2. We develop a Double-Transformer-guided Temporal
Logic framework (T2TL) that exploits the structural feature of Transformer
twice, which first encodes the LTL instruction via the Transformer
module for efficient understanding of task instructions during the
training and then encodes the context variable via the Transformer
again to capture the intrinsic relativity of sub-tasks. We evaluate our method on two continuous control
tasks. The performance and statistical analysis demonstrate the effectiveness
of our approach.

3. LTL progression, as a semantics-preserving
rewriting operation, is exploited to decompose the complex instruction
into learnable sub-goals, which not only converts non-Markovian reward
decision processes to Markovian ones, but also improves the sampling
efficiency by simultaneous learning of multiple sub-tasks. Inspired
by \cite{vaezipoor2021ltl2action}, an environment-agnostic
LTL pre-training scheme is further incorporated to facilitate the
learning of Transformer.

\section{Preliminaries\label{sec:Pre}}

\subsection{Co-Safe Linear Temporal Logic}

Co-safe LTL (sc-LTL) is a subclass of LTL that can be satisfied by
finite-horizon state trajectories \cite{Kupferman2001}. Since sc-LTL
is suitable to describe robotic instructions (e.g., trigger the alarm,
find the extinguisher, and then put out the fire), this work focuses
on sc-LTL. An sc-LTL formula is built on a set of atomic propositions
$\Pi$ that can be true or false, standard Boolean operators such
as $\wedge$ (conjunction), $\lor$ (disjunction), and $\lnot$ (negation),
temporal operators such as $\bigcirc$ (next), $\diamondsuit$ (eventually),
and $\cup$ (until). The semantics
of an sc-LTL formula are interpreted over a word $\boldsymbol{\sigma}=\sigma_{0}\sigma_{1}...\sigma_{n}$,
which is a finite sequence with $\sigma_{i}\in2^{\Pi}$, $i=0,\ldots,n$,
where $2^{\Pi}$ represents the power set of $\Pi$. Denote by $\left\langle \boldsymbol{\sigma},i\right\rangle \vDash\varphi$
if the sc-LTL formula $\varphi$ holds from position $i$ of $\boldsymbol{\sigma}$.
More detailed explanations and examples can be found in \cite{Baier2008}. 

\subsection{Labeled MDP and Reinforcement Learning}

When performing the sc-LTL task $\varphi$, the interaction between
the robot and the environment can be modeled by a labeled MDP $\mathcal{M}_{e}=\left(S,T,A,p_{e},\Pi,L,R,\gamma,\mu\right)$,
where $S$ is the state space, $T\subseteq S$
is a set of terminal states, $A$ is the action space, $p_{e}(s'|s,a)$ is the transition probability from $s\in S$
to $s'\in S$ under action $a\in A$, $\Pi$ is a set of atomic propositions
indicating the properties associated with the states, $L:S\rightarrow2^{\Pi}$
is the labeling function, $R:S\rightarrow\mathbb{R}$
is the reward function, $\gamma\in\left(0,1\right]$
is the discount factor, and $\mu$ is the initial state distribution.
The labeling function $L$ can be seen as a set of event detectors
that trigger when $p\in\Pi$ presents in the environment, allowing
the robot to determine whether or not an LTL specification is satisfied.
It is assumed that the transition probability $p_{e}$ is unknown
a priori, and the agent can only perceive its state and the
corresponding label. 

For any task $\varphi$, the robot interacts with
the environment following the policy $\pi(a|s)$ over $\mathcal{M}_{e}$. Specifically, the robot starts from
an initial state $s_{0}$ sampled from $\mu$ in each episode, and
transits from the current state $s_{t}$ to the next state $s_{t+1}$
following $p_{e}(s_{t+1}|s_{t},a_{t})$ under the control action
$a_{t}$ generated by the policy $\pi$. The robot then receives a reward by $r_{t}=R(s_{t})$. The Q-value is $Q\left(s,a\right)=\mathbb{E}\left[r_{0}+\gamma r_{1}+...|s_{0}=s,a_{0}=a,\pi\right]$
and the optimal Q-value is $Q^{*}\left(s,a\right)=\max_{\pi}Q\left(s,a\right)$.
The optimal policy $\pi^{*}$ can be derived from the optimal
Q-value. 

When applying to a large or continuous state space, the Q-value function is often parameterized with the weights function
$\theta^{Q}$ like $Q(s,a;\theta^{Q})$ in the Deep Q-Networks (DQN)
\cite{Mnih2015}. And in the continuous action case, the parameterized
policy model is often applied to the uncountable infinite problem
like $\pi_{u}(a;s,\theta^{u})$ with weights $\theta^{u}$ as in Proximal
Policy Optimization (PPO) \cite{schulman2017}. The typical reward function
is often Markovian, which means that the reward acquired at $s_{t+1}$
is only based on the transition from $s_{t}$ to $s_{t+1}$. In practice, however, the robot
is generally rewarded when the corresponding word $\boldsymbol{\sigma}$
satisfies the LTL task $\varphi$, denoted as $\boldsymbol{\sigma}\vDash\varphi$,
and the episode terminates when $\varphi$ is satisfied or falsified.
Since the word $\boldsymbol{\sigma}=\sigma_{0}\sigma_{1}...\sigma_{t}$
is formed from the state trajectory $s_{0}s_{1}...s_{t}$ through
the labeling function $L$, in this work we will
consider the non-Markovian reward function
\begin{equation}
R(s_{0}s_{1}...s_{t})=\begin{cases}
1, & \text{if }\boldsymbol{\sigma}\models\varphi\\
-1, & \text{if }\boldsymbol{\sigma}\models\lnot\varphi\\
0, & \text{otherwise}
\end{cases},\label{eq: non-Markovian reward}
\end{equation}
where $\sigma_{t}=L(s_{t})$. In the sequel, we will discuss how to
deal with the challenge of NMRDP. Given a task $\varphi$, the goal
of the agent is to learn an optimal policy $\pi^{*}(a|s)$ that maximizes
the expected discounted return $\mathbb{E}\left[\stackrel[k=0]{\infty}{\sum}\gamma^{k}r_{t+k}\mid S_{t}=s\right]$
starting from any state $s\in S$ at time step $t$.

\section{Problem Formulation\label{subsec:problem-formulation}}

To elaborate the proposed interpretable temporal logic guided reinforcement
learning algorithm, the following example will be used as a running
example throughout the work.
\begin{example}
\label{example1}
Consider a modified safety-gym environment \cite{ray2019benchmarking}, in which the robot is required
to sequentially visit a set of locations while avoiding collisions.
The set of propositions $\Pi$ is \{$\mathsf{Black\_Zone}, \mathsf{White\_Zone},\mathsf{Yellow\_Zone},\mathsf{Red\_Zone}$\}.
Using above propositions in $\Pi$, an example sc-LTL formula is $\varphi_{\mathsf{safe}}=\varphi_{\mathsf{dang}}\cup(\mathsf{Black\_Zone}\wedge(\varphi_{\mathsf{dang}}\mathsf{\cup White\_Zone}))$ where $\varphi_{\mathsf{dang}}=\neg\mathsf{Red\_Zone}\wedge\neg\mathtt{\mathsf{Yellow\_Zone}}$, which requires the robot to sequentially visit the black zone and
the white zone while avoiding colliding with red zones and yellow
zones.
\end{example}
In this work, we are interested in encoding the task conditional
states by the Transformer. By representing via Transformer
we hope to take advantage of its flexibility in encoding states and
provide interpretable analysis of the robot's motion planning. Compared
with automaton and RM-based state representations, when using Transformer
to encode the states, the gradually updated state representation can
facilitate the agent\textquoteright s comprehension of the sub-goal
at hand as the agent interacts with the environment, resulting in
a mutual improvement, in which the Transformer guides the robot\textquoteright s
motion and the selected actions improve the Transformer for better
instructions.

Specifically, suppose the representation of an LTL task $\varphi_{\theta}$
can be approximated by the Transformer parameterized with weights
$\theta_{\mathrm{trans}}$, where $\theta_{\mathrm{trans}}$ is updated
by the back-propagation of the RL controller. The goal of an interpretable
LTL guided RL in this work is to find the appropriate Transformer
weights $\theta_{\mathrm{trans}}$ over the LTL instruction, such
that an effective representation $\varphi_{\mathrm{\theta}}$ can
lead to fast learning for\textcolor{red}{{} }logical motion planning.
To this end, the problem can be formally presented as follows. 
\begin{problem}
\label{Prob1}Given a MDP $\mathcal{M}_{e}=\left(S,T,A,p_{e},\Pi,L,\gamma,\mu\right)$
corresponding to task $\varphi$ with the reward function $R_{\varphi}(s_{0}s_{1}...s_{t})$
to be designed, the goal of this work is to design an optimal representation
$\varphi_{\mathrm{\theta}}$ with $\theta_{\mathrm{trans}}^{*}$,
so that the return $\mathbb{E}\left[\stackrel[k=0]{\infty}{\sum}\gamma^{k}r_{t+k}\mid S_{t}=s\right]$
under the policy $\pi(a_{t}|s_{0}s_{1}...s_{t},\varphi)$ can be maximized.
\end{problem}

\section{Algorithm Design\label{sec:Algorithm Design}}

To address Problem \ref{Prob1}, this section presents a novel framework, namely Double-Transformer-guided
Temporal Logic framework (T2TL), that
offers interpretable LTL instruction using Transformer to guide
the robot motion planning and uses Transformer again to encode context
variables to further facilitate the robot learning. Section \ref{subsec:LTL Progression}
presents how LTL progression can be leveraged to convert NMRDP to
MDP. Section \ref{subsec:LTL Representation} explains how the Transformer
is exploited to encode the LTL specification. Section \ref{subsec:TL1}
explains in detail how Transformer can facilitate the agent's understanding
of complex tasks using simultaneous learning. Section \ref{subsec:TL2 and Pretraning}
shows how the context variable improves the agent performance and
how the pre-training scheme can be further incorporated to expedite
the convergence.

\subsection{LTL Progression and TL-MDP\label{subsec:LTL Progression}}

One of the major challenges in solving Problem \ref{Prob1} is that
the reward function $R(s_{0}s_{1}...s_{t})$ used in the Q-value function
depends on the history of the states and thus is non-Markovian. In
this work, the LTL progression from \cite{ToroIcarte2018}
is applied to solve the non-Markovian issue. Let $\mathrm{AT}(\varphi)$
denote the propositions needed to progress the current LTL specification.
The LTL progression is defined formally as follows.
\begin{defn}
\label{Def:prog}Give an LTL formula $\varphi$ and a word $\boldsymbol{\sigma}=\sigma_{0}\sigma_{1}...$,
the LTL progression $\prog\left(\sigma_{i},\varphi\right)$ at step
$i$, $\forall i=0,1,\ldots,$ is defined as follows:
\[
\begin{aligned}\prog\left(\sigma_{i},p\right) & =\mathrm{True}\text{ if }p\in\sigma_{i}\text{, where }p\in\Pi,\\
\prog\left(\sigma_{i},p\right) & =\mathrm{False}\text{ if }p\notin\sigma_{i}\text{, where }p\in\Pi,\\
\prog\left(\sigma_{i},\lnot\varphi\right) & =\lnot\prog\left(\sigma_{i},\varphi\right),\\
\prog\left(\sigma_{i},\varphi_{1}\wedge\varphi_{2}\right) & =\prog\left(\sigma_{i},\varphi_{1}\right)\wedge\prog\left(\sigma_{i},\varphi_{2}\right),\\
\prog\left(\sigma_{i},\varphi_{1}\vee\varphi_{2}\right) & =\prog\left(\sigma_{i},\varphi_{1}\right)\vee\prog\left(\sigma_{i},\varphi_{2}\right),\\
\prog\left(\sigma_{i},\ocircle\varphi\right) & =\varphi,\\
\prog\left(\sigma_{i},\varphi_{1}\cup\varphi_{2}\right) & =\prog\left(\sigma_{i},\varphi_{2}\right)\vee\left(\prog\left(\sigma_{i},\varphi_{1}\right)\wedge\varphi_{1}\cup\varphi_{2}\right).\\
\mathrm{prog}\left(\sigma_{i},\varphi\right)= & \begin{cases}
\varphi\setminus p, & \text{if }\mathrm{AT}(\varphi)=p,\text{ }\mathrm{prog}(\sigma_{i},p)=\mathrm{True},\\
\varphi, & \mathrm{otherwise}.
\end{cases}
\end{aligned}
\]
\end{defn}
The operator $\mathrm{prog}$ in Def. \ref{Def:prog} takes an LTL
formula $\varphi$ and the current label $\sigma_{i}$ as input at
each step, and outputs a formula to track which parts of the original
instructions remain to be addressed. 
\begin{thm}
\label{thm:LTL Progression}\cite{ToroIcarte2018} Given any LTL formula
$\varphi$ and the corresponding word $\boldsymbol{\sigma}=\sigma_{i}\sigma_{i+1}...$,
$\left\langle \boldsymbol{\sigma},i\right\rangle \vDash\varphi\text{ iff }\left\langle \boldsymbol{\sigma},i+1\right\rangle \vDash\mathrm{prog}\left(\sigma_{i},\varphi\right)$.
\end{thm}
There are many advantages of using the LTL progression. First, since
the operator $\prog$ can preserve LTL semantics, applying $\prog$
iteratively after each step will result in gradually diminishing LTL
instructions, which indicates the progress towards task completion.
Therefore, the reward function
can be designed by leveraging it to make the agent focus on the current
progressed task rather than the original
one all the time. Another benefit of utilizing $\prog$ iteratively
is that the complex task may be divided into a series of learnable
sub-tasks that can be viewed as simultaneous sub-goals to improve
the sampling efficiency. 
In the following context, we represent by $\Psi$ the extended training
set for which $\varphi$ and its progressed sub-tasks are included.

Based on the LTL progression in Def. \ref{Def:prog} and the LTL instruction
$\varphi$, an augmented MDP, namely the task-driven labeled MDP (TL--MDP),
is developed as follows.
\begin{defn}
\textbf{\label{Def:TL-MDP}} $\mathcal{M}_{e}=\left(S,T,A,p_{e},\Pi,L,\gamma,\mu\right)$
corresponding to an LTL task $\varphi$, the TL--MDP is constructed
by augmenting $\mathcal{M}_{e}$ to $\mathcal{M}_{\Psi}\triangleq\left\{ \left(\tilde{S},\tilde{T},A,\tilde{p},\Pi,L,\tilde{R}_{\Psi},\gamma,\mu\right):\phi_{i}\in\Psi,i=1,\ldots,\left|\Psi\right|\right\} $
with $\left|\Psi\right|$ indicating the number of tasks in $\Psi$,
where $\tilde{S}=S\times\Psi$, $\tilde{T}=\left\{ \left.(s,\phi)\right|s\in T\text{ }\textrm{or}\text{ }\phi_{i}\in\left\{ \mathrm{True},\text{ }\mathrm{False}\right\} \cup\Psi\right\} $,
$\tilde{p}((s^{'},\phi_{i}^{'})|(s,\phi_{i}),a)=p_{e}(s'|s,a)$ if
$\phi_{i}^{'}=\mathrm{prog}(L(s),\phi_{i})$ and $\tilde{p_{i}}((s^{'},\phi_{i}^{'})|(s,\phi_{i}),a)=0$
otherwise, and $\tilde{R}_{\Psi}$ is the reward function associated
with the task $\phi_{i}\in\Psi$ to overcome the non-Markovian reward
issue which can be written as
\begin{equation}
\tilde{R}_{\Psi}(s,\phi_{i})=\begin{cases}
1, & \text{if \ensuremath{\mathrm{prog}(L(s),\phi_{i})=\mathrm{True}}},\\
-1, & \text{if \ensuremath{\mathrm{prog}(L(s),\phi_{i})=\mathrm{False}}},\\
0, & otherwise.
\end{cases}\label{eq:Markovian Reward}
\end{equation}
\end{defn}
Thus, by defining the TL-MDP, the non-Markovian reward function can
be Markovian. With LTL progression, the policy $\pi_{\Psi}(a_{t}\mid s_{t},\varphi)$
that solves the LTL $\varphi$ over the TL-MDP $\mathcal{M}_{\Psi}$
can achieve the same expected discounted return as the policy $\pi_{e}(a_{t}\mid s_{0}s_{1}...s_{t},\varphi)$
in the environment $\mathcal{M}_{e}$ \cite{vaezipoor2021ltl2action}.

\subsection{Represent LTL via Transformer\label{subsec:LTL Representation}}

\begin{figure}
\centering{}\includegraphics[scale=0.26]{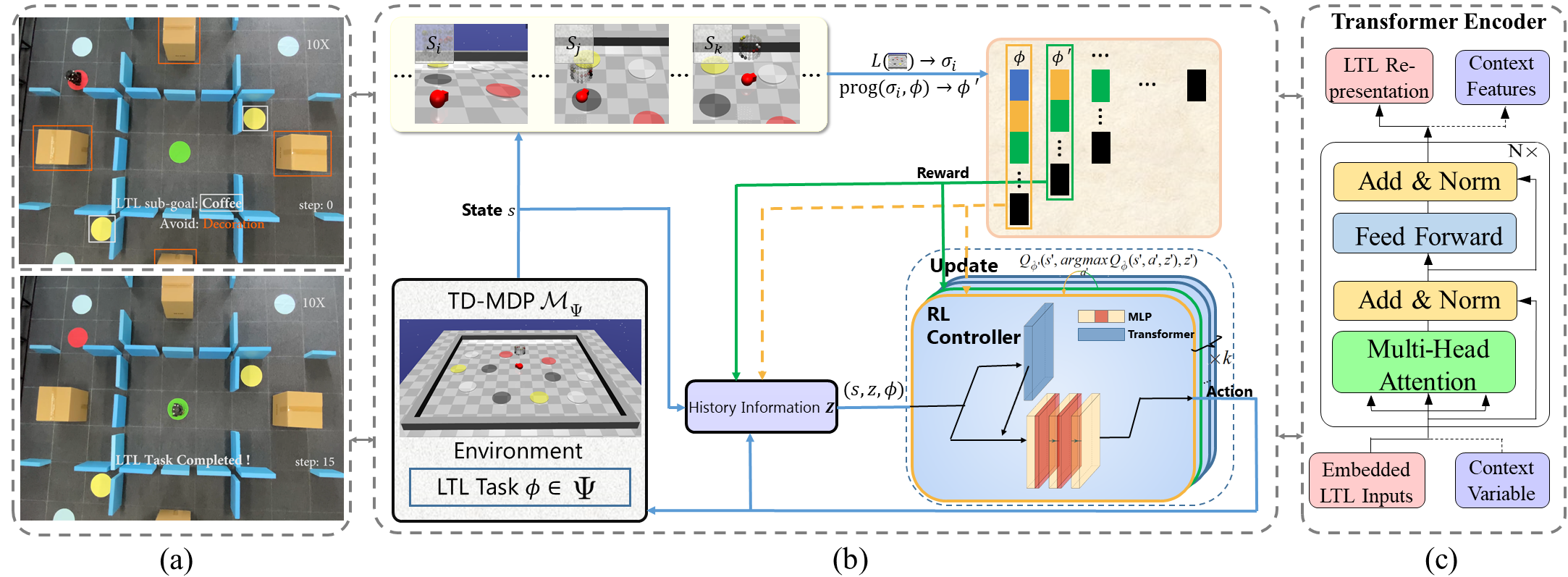}\caption{\label{fig:Framework}(a) The T2TL framework. (b) The architecture of Transformer Encoder for
T2TL framework.}
\end{figure}

Another challenge in solving Problem \ref{Prob1}
is to design an appropriate parameterized encoder for the LTL specification
for improved performance without shaping the reward function \cite{cai2023overcoming,balakrishnan2022model}
or using special exploration strategies \cite{kantaros2022} in
a sparse reward environment. To address this challenge,
inspired by the interpretable representation architecture and encoding
capability for the nature language, the Transformer from \cite{vaswani2017attention}
is exploited to represent the LTL instruction in this work. An overview
of the architecture is depicted in Fig. \ref{fig:Framework}(b). 

\begin{figure}
\centering{}\includegraphics[scale=0.23]{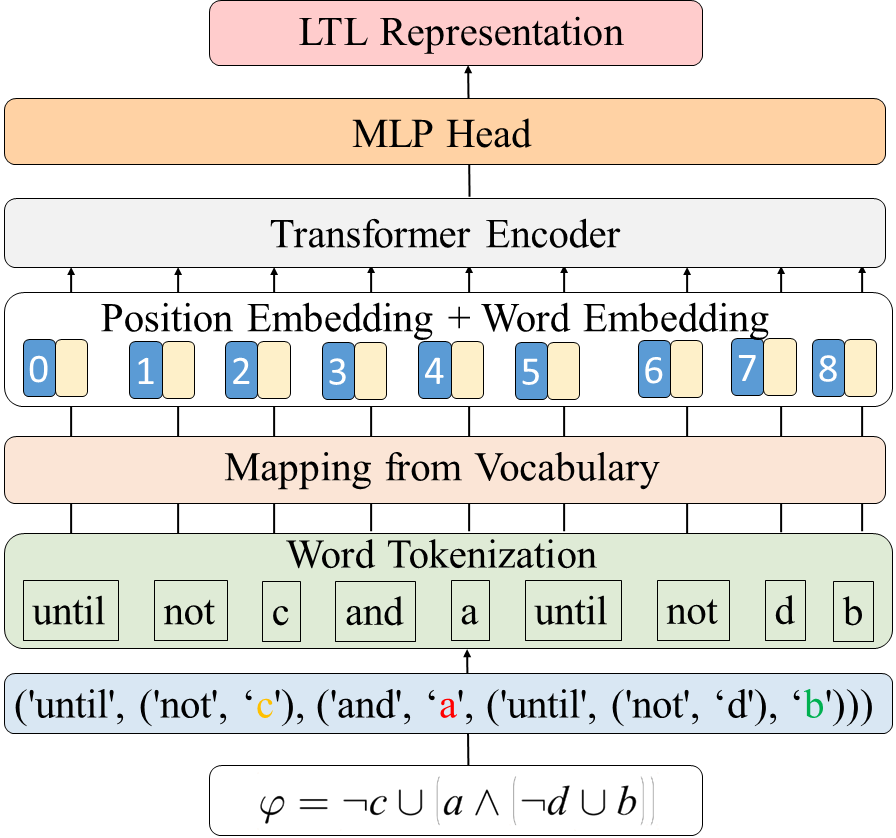}\caption{\label{fig:LTL2Repre}An example of the LTL Representation encoded
via Transformer.}
\end{figure}
Given an input $X_{\varphi}=(x_{0},x_{1},...)$ generated by the LTL
task $\varphi$ where $x_{t},t=0,1,...,$ represents the operator
or proposition, $X_{\varphi}$ will be preprocessed by the word embedding
$\mathrm{E}$ as $X_{\mathrm{E}}=\left[x_{0}\mathrm{E};x_{1}\mathrm{E};\ldots;x_{N}\mathrm{E}\right]\in\mathbb{R}^{B\times(N+1)\times D}$
where $B$ is the batch size, $N+1$ is the length of input $X_{\varphi}$,
and $D$ is the model dimension of the Transformer. $X_{\mathrm{E}}$
is then added with the frequency-based positional embedding 
$E_{pos}$ to make use of the order of the sequence. For instance,
a task $\varphi=\lnot\mathrm{c}\cup\left(\mathrm{a}\wedge\left(\lnot\mathrm{d}\cup\mathrm{b}\right)\right)$
can be encoded as shown in Fig.
\ref{fig:LTL2Repre}.

The encoder is constructed by stacking identical transformer layers
and each transformer layer is built with a self-attention sub-layer
and a position-wise fully connected feed-forward (MLP) sub-layer.
Layer norm (LN) is applied before every sub-layer and residual connections
are applied after every block. In the structure of the Transformer,
the multi-head self-attention (MSA) method plays an important role
in establishing the intrinsic connections between words. Specifically,
given the query $Q$, key $K$, and value $V$ derived from the LTL
input $X_{\varphi}=(x_{0},x_{1},...)$, the similarity of words can
be calculated by the dot-product attention as
\[
Attention(Q,K,V)=softmax\left(\frac{QK^{T}}{\sqrt{d_{k}}}\right)V,
\]
where $\sqrt{d_{k}}$ is the scaling factor. The global computation
procedure of the encoder layers is represented as follows: 
\[
\begin{array}{cc}
X_{0}=\left[x_{0}\mathrm{E};x_{1}\mathrm{E};...;x_{N}\mathrm{E}\right]+E_{pos}, & E_{pos}\in\mathbb{R}^{B\times(N+1)\times D}\\
X_{l}^{'}=\text{MSA}(\text{LN}(X_{l-1}))+X_{l-1}, & l=1,...,L\\
X_{l}=\text{MLP}(\text{LN}(X_{l}^{'}))+X_{l}^{'}, & l=1,...,L\\
Y=\text{LN}(X_{l})
\end{array}
\]
where $Y$ represents the output of the last layer from the Transformer
encoder, which can be manually customized to an appropriate dimension
according to the need of tasks.

Motivated by \cite{vig2020investigating} and \cite{Vig19}, the weights or heads of self-attention in Transformer can offer reasonable interpretability for the agent\textquoteright s motion planning in RL. Specifically, given the weights $W_{H}^{L}\in\mathbb{R}^{L\times H\times (N+1)\times (N+1)}$, the interpretability can be indicated by showing on which proposition (i.e., the sub-task in LTL to be solved) the head\textquoteright s weights are more focused according to $token^{*}=\underset{m\in M}{\arg\max}\stackrel[l=0]{L}{\sum}\stackrel[h=0]{H}{\sum}W_{h,m}^{l},$ where $M=\{1,2,...,N+1\}$ and $H$ is the number of heads in
Transformer. Note that, since this work involves Transformer inputs
that do not consider a co-reference candidate, (e.g., the
gender bias), all heads are equally important and do not have pre-set
emphasis tokens to generate top heads like \cite{vig2020investigating}.

\subsection{T1TL and Simultaneous Learning \label{subsec:TL1}}

As shown in Table \ref{tab:Comparison}, traditional product-MDP
algorithms usually represent the states of automaton or RM with one-hot
encoding or sorted index, whose representation
needs to be customized manually and the dimensions are dependent on the complexity of the LTL task. Unlike these works, we encode the
LTL specification as normalized vectors using Transformer, which is
not only appropriate for the forward propagation of the neural network,
but also can be continuously updated as Transformer evolves. In addition,
its dimension can be customized with appropriate designs of Transformer,
leading to improved agent\textquoteright s performance. Compared with
automaton-based methods, the product-MDP based Transformer can be
constructed on-the-fly without concern of exponential explosion of
algorithm complexity with LTL tasks. 

\begin{table}
\caption{\label{tab:Comparison}The Comparison of LTL Representations between
Traditional Methods and Transformer}

\centering{}\resizebox{0.48\textwidth}{!}{
\begin{tabular}{c|c|c}
\hline 
 & \multicolumn{1}{c|}{Automaton or RM} & \multicolumn{1}{c}{Transformer}\tabularnewline
\hline 
Dimension & fixed (limited by LTL task complexity) & flexible\tabularnewline
\hline 
Representation & customize manually & update via Transformer\tabularnewline
\hline 
Construction & in advance (in most cases) & on-the-fly\tabularnewline
\hline 
\multirow{2}{*}{Interpretability} & \multicolumn{1}{c|}{indirect (interpreted by some} & direct (interpreted by weights\tabularnewline
 & module in other models) \cite{zhang2021decision,araki2021learning,li2021learning} & or heads in self-attention)\tabularnewline
\hline 
Effect & limited by dimension & better with appropriate dimensions\tabularnewline
\hline 
\end{tabular}}
\end{table}
\begin{figure}
\centering{}\includegraphics[scale=0.27]{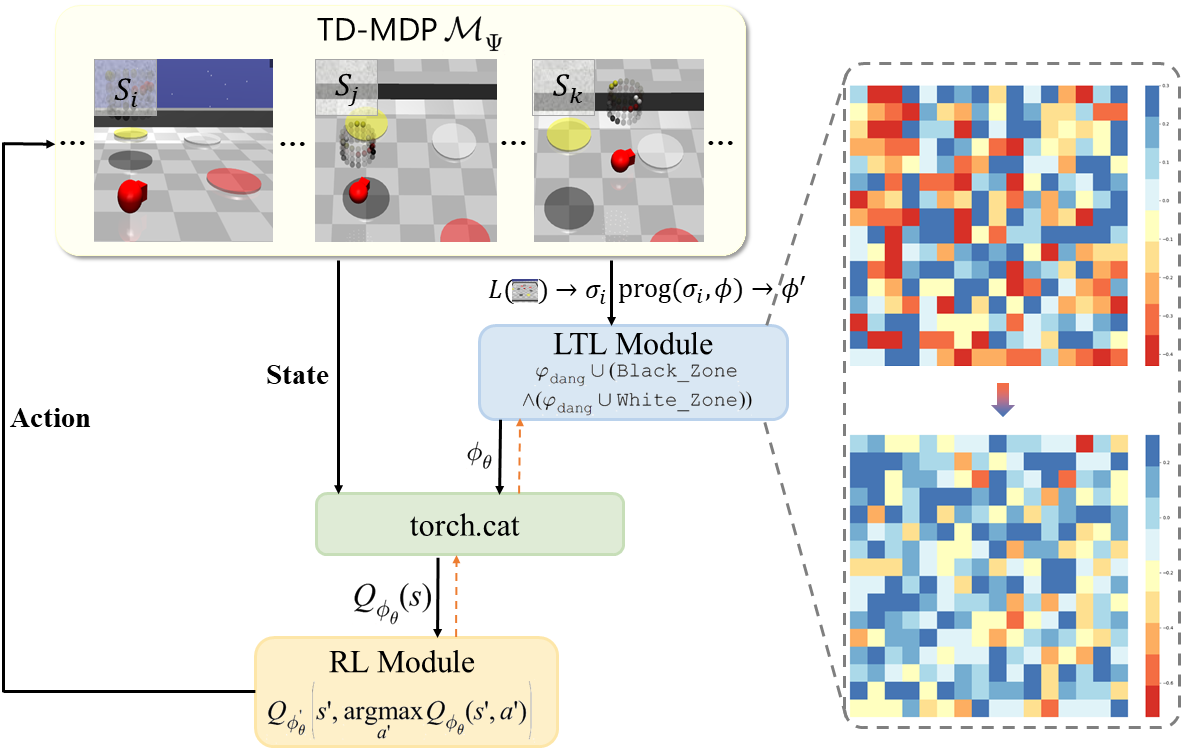}\caption{\label{fig:BP4TL1}The outline of the Transformer module updated via
interactions between the agent and the environment.
The heatmap depicts the update process of the self-attention out-projection
weights in Transformer from the state $(s_{i},\phi)$ to state$(s_{k},\phi^{'}).$
The orange dashed line shows the back-propagation of the LTL representation
encoded via Transformer. }
\end{figure}
The interpretable LTL representation encoded via the Transformer
is illustrated in Fig. \ref{fig:BP4TL1}. Initially, the weights of Transformer
are set randomly. As the agent interacts with the environment, the
RL module is updated when a proposition is encountered by the agent,
which leads to an indirect update of the Transformer
module, i.e., the agent has new knowledge of the pros and cons about
the currently encountered proposition for completing the task. Thus,
as the RL module converges, the Transformer module achieves a better
representation of the LTL instruction. Meanwhile, as the representation
of LTL becomes more effective, the convergence of the RL policy is
further improved. Let $Q_{\varphi_{\theta}}\left(s,a\right)$ and
$Q_{\varphi_{\theta}^{'}}\left(s,a\right)$ be the Q-value function
of task $\varphi$ and $\varphi^{'}$, respectively. Thus in the conventional RL algorithm, such as DQN,
the update for $Q_{\varphi_{\theta}}$ driven by the Transformer can be written as 
\[
Q_{\varphi_{\theta}}\leftarrow Q_{\varphi_{\theta}}+\alpha\left(R_{\varphi}+\gamma\underset{a'}{\mathrm{max}}Q_{\varphi_{\theta}^{'}}\left(s',a'\right)-Q_{\varphi_{\theta}}\right).
\]
However, conventional off-policy DRL algorithm usually performs a random exploration
in the early stage. If an action effective
for other tasks is performed rather than the current task, such an
action is often ignored and will not be utilized to update the Q-value
for associated tasks, resulting in low sampling
efficiency and delayed convergence to the optimal policy. Note that the on-policy DRL algorithms, such
as PPO, usually train the agent with parallel environments to improve
sampling efficiency by reducing the correlation of transition data.
However, this trick usually can't be applied to off-policy DRL algorithm
due to the experience replay buffer.

Compared with vanilla DQN, the idea of simultaneous learning
is to extract sub-tasks from $\varphi$ via LTL progression as described
in Sec. \ref{subsec:LTL Progression}, augment the original MDP $\mathcal{M}_{e}$
with the LTL representation encoded by Transformer module,
and use Q-learning to simultaneously learn these sub-tasks. 

Particularly, the simultaneous learning begins with extracting sub-tasks
from $\varphi$ by LTL progression to generate an extended training
set $\Psi$. All tasks $\phi\in\Psi$
are associated with a Q-value function $Q_{\phi_{\theta}}\left(s,a\right)$
where the LTL instruction is encoded by the Transformer module, and
a series of episodes over the tasks in $\Psi$ is performed using
the off-policy learning method. For each $\phi\in\Psi$,
the robot updates the Q-value functions as if it is currently trying
to solve $\phi$. Specially, given the current state $s$, the formula
$\phi^{'}$ will be the progressed LTL task if $\phi^{'}=\mathrm{prog}\left(L\left(s\right),\phi\right)$.
The robot selects an action $a$ following a behavior policy (e.g.,
the $\varepsilon$-greedy one) based on the Q-value $Q_{\phi_{\theta}}$
and then transits to the next state with rewards received from (\ref{eq:Markovian Reward}).\textcolor{brown}{{}
}Let $Q_{\phi_{\theta}}\left(s,a\right)$ and $Q_{\phi_{\theta}^{'}}\left(s,a\right)$
be the Q-value function of task $\phi$ and $\phi^{'}$, respectively.\textcolor{brown}{{}
}Thus under the simultaneous learning, $Q_{\phi_{\theta}}$
is updated following a modified double DQN as
\begin{equation}
\begin{alignedat}{1}Q_{\phi_{\theta}}\leftarrow & Q_{\phi_{\theta}}+\alpha\left(\tilde{R}_{\Psi}+\right.\\
 & \left.\gamma Q_{\phi_{\theta}^{'}}\left(s',\underset{a'}{\mathrm{argmax}}Q_{\phi_{\theta}}(s',a')\right)-Q_{\phi_{\theta}}\right)
\end{alignedat}
.\label{eq: Update based on TL1}
\end{equation}
By this way, the Q-value of $\phi$ will be propagated
backwards from its sub-tasks $\phi^{'}$ and the weights of Transformer
will also be updated over the state representation. Thus by developing
TL-MDP $\mathcal{M}_{\Psi}$, it will not only convert
the non-Markovian reward processes to Markovian ones, but also provide
simultaneous update for sub-task's Q-value. Such
a method enables the update of the current $Q_{\phi_{\theta}}$
and its sub-task $Q_{\phi_{\theta}^{'}}$, resulting in an effective
representation of LTL for improved convergence. 

\subsection{T2TL and Pre-training Scheme\label{subsec:TL2 and Pretraning}}

\begin{algorithm}
\caption{\label{Alg2}T2TL with Pre-training Scheme}

\scriptsize

\singlespacing

\begin{algorithmic}[1]

\Procedure {Input:} {An LTL intruction $\varphi$ and the MDP
$\mathcal{M}_{e}$ corresponding to $\varphi$}

{Output: } { An approximately optimal stationary policy $\pi_{\Psi}^{*}(a_{t}\mid s_{t},\varphi)$
for the TL-MDP $\mathcal{M}_{\Psi}$ }

{Initialization: } { All neural network weights }

\State Load the pre-trained weights to the Transformer module, extract
sub-tasks as $\Psi$, and initialize $Q_{\phi_{\theta}}$ and $Q_{\phi_{\theta}^{'}}$
for $\phi$ and its sub-task $\phi^{'}$

\While {\textbf{ $T<T_{max}$} }

\State Augment the state $s$ with $\phi_{\theta}$ encoded by Transformer,
and set the context variable to zero

\While {\textbf{ $t<t_{max}$} }

\State $\phi^{'}\leftarrow\mathrm{prog}(L(s),\phi)$

\If {$\phi^{'}\in\{\mathsf{True,\mathsf{False}}\}$ or $s\in T$
}

\State Break

\EndIf

\State Gather data from $\phi$ and encode the context variable through Transformer

\For {$Q_{\phi_{\theta}}\in Q$}

\State $\phi^{'}\leftarrow\mathrm{prog}(L(s),\phi)$

\State Determine $\tilde{R}_{\tilde{\phi}}$ by (\ref{eq:Markovian Reward})
and update $Q_{\phi_{\theta}}$ following (\ref{eq:TL2_update})

\EndFor

\State $t\leftarrow t+1$

\EndWhile

\State $T\leftarrow T+1$

\EndWhile

\EndProcedure

\end{algorithmic}
\end{algorithm}
Since LTL progression decomposes the original LTL specification into
sub-goals that can be learned simultaneously in Sec. \ref{subsec:TL1},
the context variable that captures the connections of simultaneous
sub-goals is further incorporated using Transformer. The context variable in meta reinforcement learning (meta-RL) \cite{Wang2016}
is used to capture the intrinsic relativity of multiple
tasks. In \cite{Rakelly2019}, an off-policy meta-RL with the
probabilistic context variable is developed, which enhances adaptation
efficiency using posterior sampling during training. The work of \cite{FakoorCSS20} adopts the deterministic context
variable to further improve the learning performance.

Considering the effectiveness and compatibility, the deterministic
context variable $Z$ is applied. Specially, a deterministic context variable $z\in Z$
acts as a fixed length window and extracts the knowledge of history
observations, actions and rewards in a certain range when the agent
explores the environment. Different from \cite{FakoorCSS20} that
uses RNN, Transformer is leveraged to encode the context variable
in this work, which facilitates the convergence of the LTL representation. Thus the Q-value
function $Q_{\phi_{\theta}}(s,a)$ is then conditioned on the context
as $Q_{\phi_{\theta}}(s,a,z)$, where $z\in Z$ is a deterministic
context variable, and (\ref{eq: Update based on TL1}) can be augmented
as 
\begin{equation}
\begin{alignedat}{1}Q_{\phi_{\theta}}\leftarrow & Q_{\phi_{\theta}}+\alpha\left(\tilde{R}_{\Psi}+\right.\\
 & \left.\gamma Q_{\phi_{\theta}^{'}}\left(s',\underset{a'}{\mathrm{argmax}}Q_{\phi_{\theta}}(s',a',z'),z'\right)-Q_{\phi_{\theta}}\right)
\end{alignedat}
.\label{eq:TL2_update}
\end{equation}
By this way, the robot is able to comprehend the LTL task by considering
the context information and expedite the learning in a sparse reward
environment.

Inspired by the competitive performance on downstream tasks when using
the pre-training method in \cite{vaezipoor2021ltl2action},
an environment-agnostic module is further incorporated as the pre-training
scheme in this work. First, a single-state MDP $\mathcal{M}_{s}=\left(S,T,A,p_{s},\Pi,L,\gamma,\mu\right)$
is built, where $S=\{s_{o}\}$, $T=\emptyset$, $A=\Pi$, $p_{s}(s_{0}\mid s_{0},\cdot)=1$,
$\mu(s_{0})=1$ and $L(s_{0})=\{p\}$. Then the single-state MDP $\mathcal{M}_{s}$
can be augmented to TL-MDP $\mathcal{M}_{\Psi_{s}}$ with the LTL
instruction $\varphi$. Second, the agent tries to complete the LTL
task in each episode until the Transformer module converges. At the end of the pre-training,
the learned Transformer weights are then transferred to the downstream
MDP as the initial LTL Module (e.g., the TL-MDP $\mathcal{M}_{\Psi}$
in the revised safety-gym). Note that the design
of $A=\Pi$ is to learn a policy that satisfies the LTL task as quickly
as possible by choosing one proposition to be true at each time step.
With the pre-training scheme, the LTL presentation
from $\varphi$ can help the agent infer which part of the information
should be emphasized to increase the probability of achieving sub-goals. The overall method is illustrated
in Fig. \ref{fig:Framework}(a) and the pseudo-code is outlined in
Alg. \ref{Alg2}.

\section{CASE STUDIES}

In this section, the developed T2TL framework is evaluated against
the state-of-the-art algorithms in simulation\footnote{Our codes are avaliable at https://github.com/Charlie0257/T2TL}.
Specifically, we consider the following aspects. \textbf{1) Performance:}
how well does our approach outperform the state-of-the-art algorithms
in two continuous environments? \textbf{2) Representation:}
What is the role of the representation dimensions for LTL specifications?\textbf{
3) Interpretability:} How well can the agent understand LTL specifications
via Transformer? 

To show the effectiveness of the T2TL framework, denoted by $\mathrm{T2TL}_{\mathrm{pre}}$,
it is empirically compared with four baselines. The first baseline
is DFA from \cite{Lacerda2014} which is used to construct the product
MDP for the LTL task over a finite horizon. The second baseline is
RM from \cite{Icarte2022} which has automaton-based representations
that exploit the reward function\textquoteright s internal structure
to learn optimal policies. The third baseline is $\mathrm{GNN}_{\mathrm{pre}}$,
which uses a pre-training scheme from \cite{vaezipoor2021ltl2action} and exploits
the compositional syntax and semantics of LTL by GNN to solve complex
multiple tasks. Note that the simultaneous learning is incorporated
in $\mathrm{GNN}_{\mathrm{pre}}$ for fair comparisons with our method.
The fourth baseline is $\mathrm{T1TL}_{\mathrm{pre}}$,
which exploits Transformer instead of GNN to
encode LTL instructions with a pre-training scheme. The fifth baseline is $\mathrm{DFA}_{\mathrm{cont}}$,
which is based on DFA and uses Transformer to encode the context variable
to capture the intrinsic connection between sub-tasks.

To evaluate the performance in a sparse reward environment, our framework
is verified in two different continuous cases. The RL algorithms
applied to two cases are double DQN \cite{VanHasselt2016} and PPO
\cite{schulman2017} respectively to show the generality of our method
over the on-policy and off-policy RL.

\begin{figure}
\centering{}\includegraphics[scale=0.28]{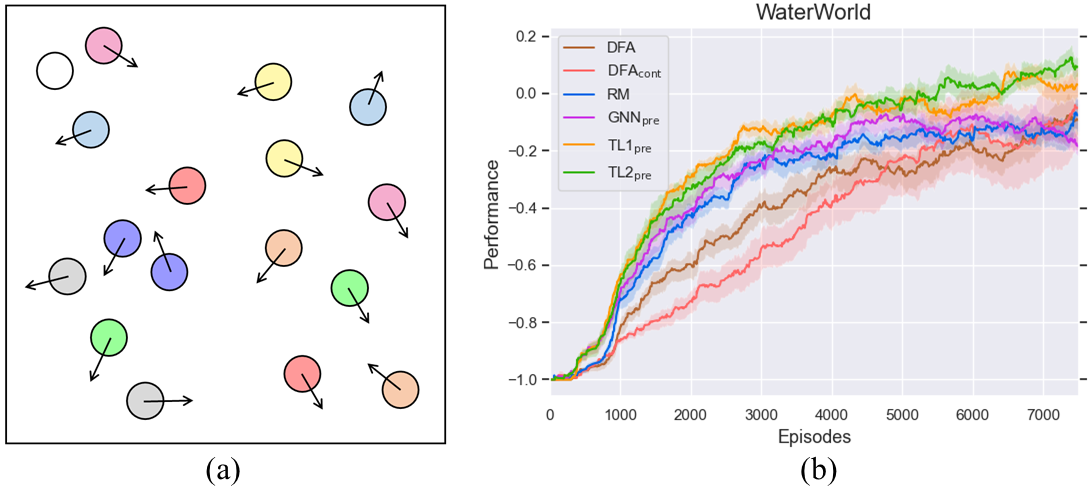}\caption{\label{fig:Water World} (a) The WaterWorld environment. (b) The performance
of different methods in the WaterWorld scenario.}

\end{figure}
\textbf{(1) Case 1: WaterWorld.}{}
We first evaluate the developed T2TL framework in a dynamic continuous
world \cite{Icarte2022}. As shown in Fig. \ref{fig:Water World}(a),
each ball moves at a fixed velocity in a certain direction and bounces
when it hits a wall. The agent represented by the white ball can increase
its speed in any of the four cardinal directions. The set of propositions
$\Pi$ in this environment is composed of balls of different colors.
In this scenario, we consider an sc-LTL task $\varphi_{\mathsf{water}}=\varphi_{\mathsf{avoid}}\cup(\mathsf{Yellow}\wedge(\varphi_{\mathsf{avoid}}\cup(\mathsf{Purple}\wedge(\varphi_{\mathsf{avoid}}\cup(\mathsf{Megenta}\wedge(\varphi_{\mathsf{avoid}}\cup(\mathsf{Orange}\wedge(\varphi_{\mathsf{avoid}}\cup\mathsf{Gray}))))))))$,
where $\varphi_{\mathsf{avoid}}=\lnot\mathsf{Pink}\wedge\lnot\mathsf{Green}\wedge\lnot\mathsf{Blue}$,
which requires the agent to encounter the ball with $\mathsf{Yellow}$,
$\mathsf{Purple}$, $\mathsf{Megenta}$, $\mathsf{Orange}$ and $\mathsf{Gray}$
in order while avoiding $\mathsf{Pink}$, $\mathsf{Green}$ and $\mathsf{Blue}$
balls.

Fig. \ref{fig:Water World}(b) shows the performances
of all baselines against ours over the task $\varphi_{\mathsf{water}}$
with 12 random seeds in the WaterWorld environment. Clearly, the method of
simultaneous learning shows improved convergence than $\mathrm{DFA}$.
By encoding the LTL representation using Transformer, $\mathrm{T1TL}_{\mathrm{pre}}$
outperforms RM and $\mathrm{GNN}_{\mathrm{pre}}$. By
incorporating context variable, $\mathrm{T2TL}_{\mathrm{pre}}$ shows better performance at the
end.

\begin{figure}
\centering{}\includegraphics[scale=0.23]{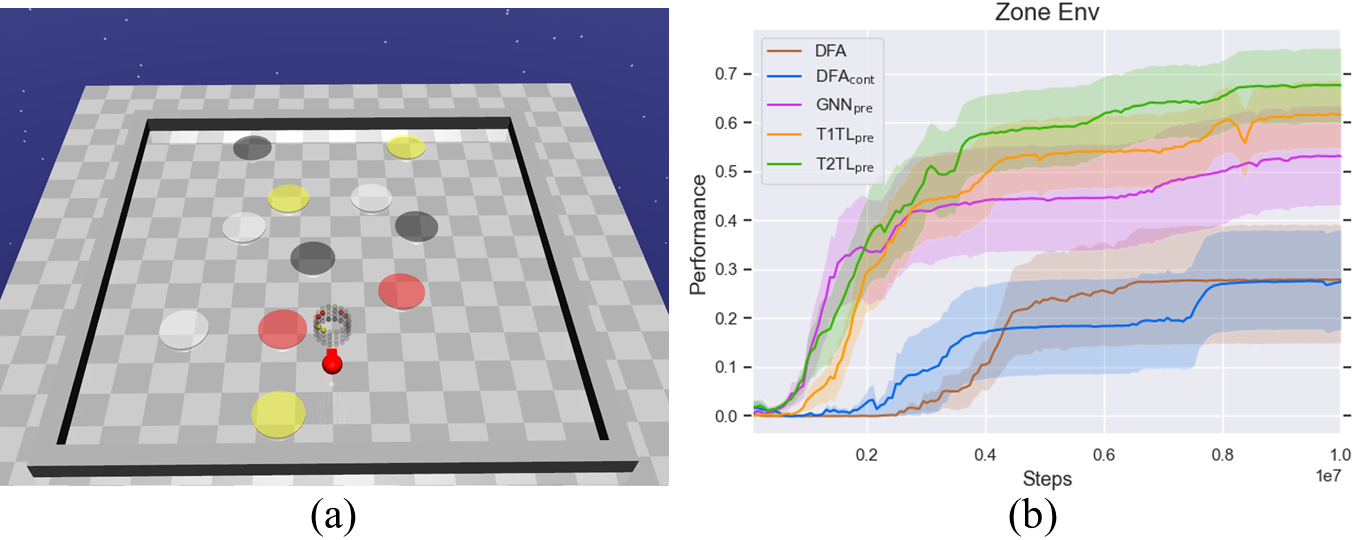}\caption{\label{fig:safety case} (a) The ZoneEnv case.
(b) The performance of different methods for the ZoneEnv scenario.}
\end{figure}
 \textbf{(2) Case 2: ZoneEnv. }We further evaluate our framework in
a modified Safety-gym \cite{ray2019benchmarking} environment as shown
in Fig. \ref{fig:safety case}(a). Consider a sequential task requiring
the robot to visit the red zone, black zone, and yellow zone in order,
which can be written as $\varphi_{\mathsf{zone}}=\lozenge(\mathsf{Red\_Zone}\wedge\lozenge(\mathsf{Black\_Zone}\wedge\lozenge\mathsf{Yellow\_Zone}).$ Fig. \ref{fig:safety case}(b)
shows the performance of $\mathrm{GNN}_{\mathrm{pre}}$ and $\mathrm{T1TL}_{\mathrm{pre}}$,
which outperform $\mathrm{DFA}$ clearly,
reflecting the effect of the LTL representation encoded by neural networks. $\mathrm{T2TL_{pre}}$ shows a competitive
performance compared with $\mathrm{T1TL_{pre}}$ and $\mathrm{GNN}_{\mathrm{pre}}$.

\begin{table}
\caption{\label{tab:task_5+n}The steps statistics over the five sub-goals
with increasing obstacles between different representation methods.}

\centering{}\resizebox{0.47\textwidth}{!}{
\begin{tabular}{c|ccc}
\hline 
Task Level & T1TL\_pre & GNN\_pre & RM\tabularnewline
\hline 
5 sub-goals with 1 obstacle & \textbf{236.50($\boldsymbol{\pm}$4.49)} & 253.26($\pm$6.57) & 259.13($\pm$3.06)\tabularnewline
\hline 
5 sub-goals with 2 obstacles & \textbf{286.77($\boldsymbol{\pm}$5.21)} & 310.50($\pm$6.24) & 322.56($\pm$13.47)\tabularnewline
\hline 
5 sub-goals with 3 obstacles & \textbf{421.31($\boldsymbol{\pm}$8.32)} & 468.73($\pm$9.02) & 489.07($\pm$14.17)\tabularnewline
\hline 
\end{tabular}}
\end{table}
 
\begin{table}
\caption{\label{tab:task_n+1}The steps statistics over the increasing sub-goals
with one obstacle between different representation methods.}

\centering{}\resizebox{0.47\textwidth}{!}{
\begin{tabular}{c|ccc}
\hline 
Task Level & T1TL\_pre & GNN\_pre & RM\tabularnewline
\hline 
4 sub-goals with 1 obstacle & \textbf{189.46($\boldsymbol{\pm}$1.98)} & 191.71($\pm$3.25) & 207.21($\pm$8.03)\tabularnewline
\hline 
5 sub-goals with 1 obstacle & \textbf{236.50($\boldsymbol{\pm}$4.49)} & 253.26($\pm$6.57) & 259.13($\pm$3.06)\tabularnewline
\hline 
6 sub-goals with 1 obstacle & \textbf{609.82($\boldsymbol{\pm}$15.73)} & 628.29($\pm$21.92) & 659.32($\pm$6.39)\tabularnewline
\hline 
\end{tabular}}
\end{table}
\textbf{ }
\begin{table}
\caption{\label{tab:task_5+2-n}The steps statistics over the five sub-goals
with two obstacles in increasing size between different representation
methods.}

\centering{}\resizebox{0.47\textwidth}{!}{
\begin{tabular}{c|ccc}
\hline 
Task Level: & T1TL\_pre & GNN\_pre & RM\tabularnewline
\hline 
5 sub-goals with 2 obstacles (size: 1x) & \textbf{286.77($\boldsymbol{\pm}$5.21)} & 310.50($\pm$6.24) & 322.56($\pm$9.47)\tabularnewline
\hline 
5 sub-goals with 2 obstacles (size: 1.5x) & \textbf{291.94($\boldsymbol{\pm}$2.18)} & 305.85($\pm$8.04) & 338.57($\pm$10.24)\tabularnewline
\hline 
5 sub-goals with 2 obstacles (size: 2.5x) & \textbf{269.88($\boldsymbol{\pm}$5.00)} & 276.34($\pm$7.48) & 300.78($\pm$12.20)\tabularnewline
\hline 
\end{tabular}}
\end{table}
\textbf{(3) Statistical Analysis for Representations.
}To further show the benefits of Transformer-encoded representation, we compare the average steps used
by different representation methods when completing the task of different complexities in the WaterWorld environment. As shown in Table
\ref{tab:task_5+n}, as the number of obstacles increases, the Transformer-encoded representation uses  fewer steps than the baselines. In Table \ref{tab:task_n+1}, the representation encoded by
Transformer still shows better performance than the baselines. Table \ref{tab:task_5+2-n}
shows more stable performance can be achieved using Transformer when the size of the obstacle becomes larger.

\begin{table}
\caption{\label{tab:Unit Performance for WaterWorld} The Performance of Unit
Time Statistics Over All Algorithms in the WaterWorld Scenario (min). }
\centering{}\resizebox{0.47\textwidth}{!}{
\begin{tabular}{c|ccccc}
\hline 
WaterWorld & $\mathrm{T2TL}_{\mathrm{pre}}$ & $\mathrm{T1TL}_{\mathrm{pre}}$ & $\mathrm{GNN}_{\mathrm{pre}}$ & RM & $\mathrm{DFA}$\tabularnewline
\hline 
$Perf_{Unit}$ & 1.96($\pm$0.08) & \textbf{2.60($\pm$0.05)} & 1.95($\pm$0.03) & 2.20($\pm$0.20) & 2.02($\boldsymbol{\pm}$0.52)\tabularnewline
\hline 
\end{tabular}}
\end{table}
 
\begin{table}
\caption{\label{tab:Unit Performance for ZoneEnv} The Performance of Unit
Time Statistics Over All Algorithms in the ZoneEnv Scenario ($\times10^{-2}$/min). }
\centering{}\resizebox{0.47\textwidth}{!}{
\begin{tabular}{c|ccccc}
\hline 
ZoneEnv & $\mathrm{T2TL}_{\mathrm{pre}}$ & $\mathrm{T1TL}_{\mathrm{pre}}$ & $\mathrm{GNN}_{\mathrm{pre}}$ & $\mathrm{DFA}_{\mathrm{cont}}$ & $\mathrm{DFA}$\tabularnewline
\hline 
$Perf_{Unit}$ & \textbf{12.21($\pm$7.16)} & 11.09($\pm$7.46) & 9.60($\pm$5.85) & 3.12($\pm$2.80) & 6.61($\boldsymbol{\pm}$5.63)\tabularnewline
\hline 
\end{tabular}}
\end{table}
 \textbf{(4) Statistical Analysis for Performance
in Unit Time. }To further evaluate the performance
of different methods per unit of time, we define the unit performance as $Perf_{Unit}=\frac{\sideset{}{_{i=1}^{N}}\sum R_{i}}{\sideset{}{_{i=1}^{N}}\sum T_{i}+T_{pre}},$
where $N$ is the number of
total episodes or total steps, $R_{i}$ is the reward in one episode
or fixed steps, $T_{i}$ is the elapsed time in one episode or fixed
steps, and $T_{pre}$ is the pre-training time. As shown in Table \ref{tab:Unit Performance for WaterWorld},
$\mathrm{T2TL}_{\mathrm{pre}}$ spends more time updating
the double Transformer framework, while  $\mathrm{T1TL}_{\mathrm{pre}}$
yields better performance
compared to other algorithms. As shown in Table \ref{tab:Unit Performance for ZoneEnv},
$\mathrm{T2TL}_{\mathrm{pre}}$ can improve the sampling efficiency and shows better performance in more challenging environments.

\begin{figure}
\centering{}\includegraphics[scale=0.25]{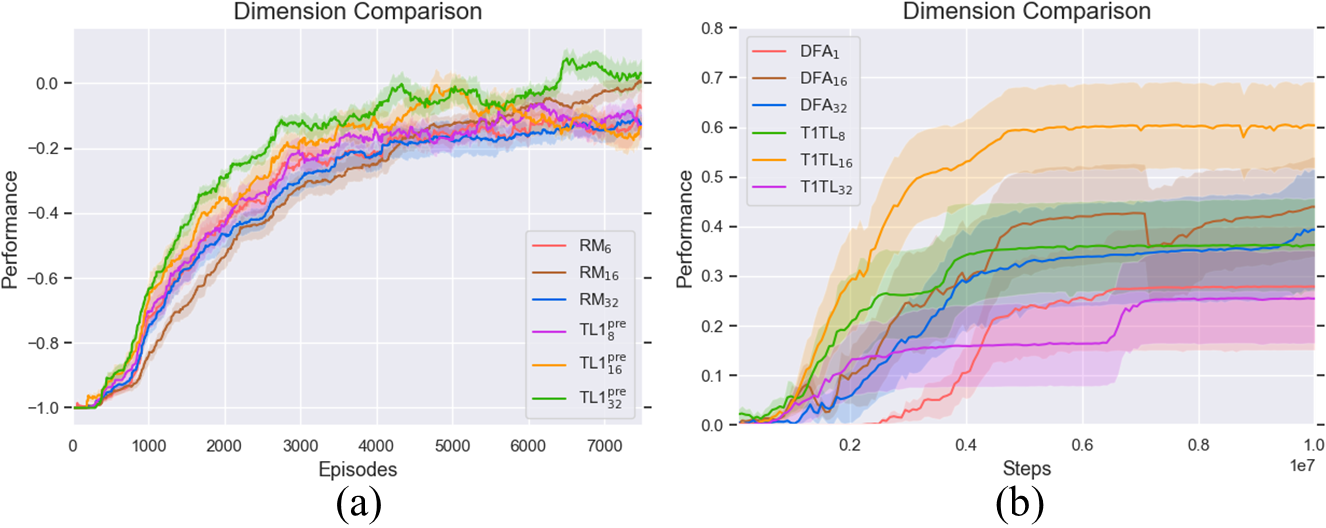}\caption{\label{fig:Dimension Comparison} (a) The performance between
$\mathrm{T1TL}_{\mathrm{pre}}$ and RM with different representation
dimensions in the WaterWorld case. (b) The
performance between $\mathrm{T1TL}$ and DFA with different
representation dimensions in the ZoneEnv case.}
\end{figure}
 \textbf{(5) Dimension Comparison.} To emphasize the influence of
the representation dimension of the LTL instruction on agent performance
in a high-dimensional state space or complex environment, Fig.
\ref{fig:Dimension Comparison} shows the results between $\mathrm{T1TL}_{\mathrm{pre}}$
and traditional methods by {one-hot
encoding with different dimensions in the WaterWorld and ZoneEnv environment.  It is clear in Fig. \ref{fig:Dimension Comparison}(a)
that an appropriate increase for the representational dimension of
the LTL instruction is beneficial in providing agents
with more comprehensive information. In Fig. \ref{fig:Dimension Comparison}(b),
an appropriate representational dimension further improves the performance
of DFA, and the agent achieves good performance when the dimension
is 16 in $\mathrm{T1TL}$.

\begin{figure}
\centering{}\includegraphics[scale=0.24]{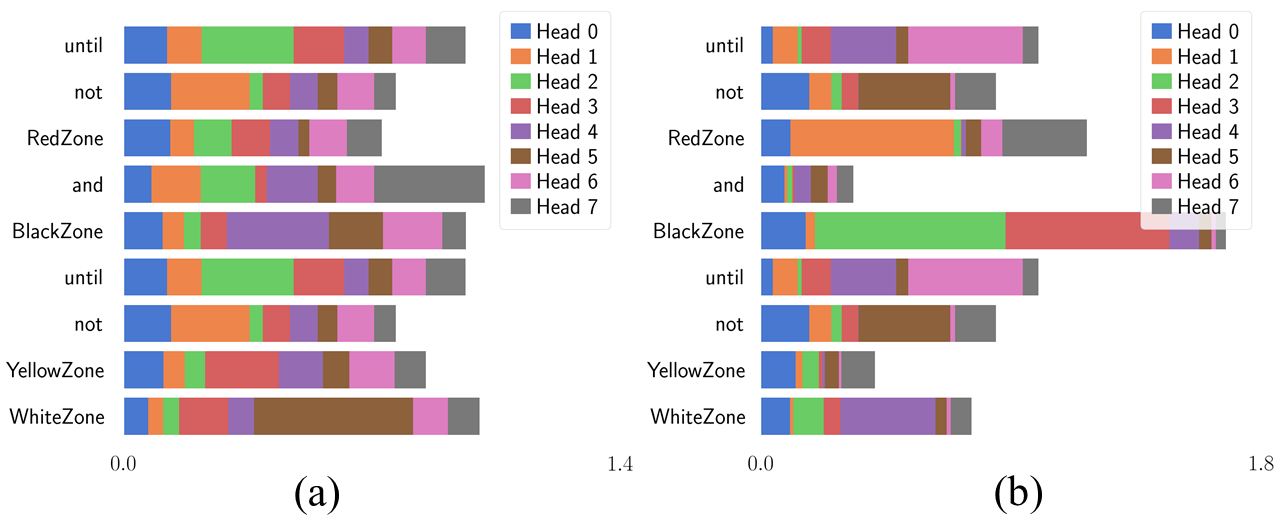}\caption{\label{fig:Interpretable} The heads concentration from attention
view of Transformer on the instruction\textcolor{red}{{} }$\varphi_{\mathsf{zone'}}$.
(a) and (b) reflect the process of the comprehension of the agent
to the LTL instruction.}
\end{figure}
\textbf{(6) Interpretability via Attention.} To further visualize
how well the agent understand the LTL task $\varphi_{\mathsf{zone'}}=\lnot\mathsf{Red\_Zone}\cup(\mathsf{Black\_Zone}\wedge(\lnot\mathsf{Yellow\_Zone}\cup\mathsf{White\_Zone}))$
when the Transformer module converges, Fig. \ref{fig:Interpretable} shows a view from heads in
attention to interpret which tokens the agent would be more interested
in. In Fig. \ref{fig:Interpretable}, different color bars represent
different heads in the layers of attention and its length indicates
the weights of the head on this token. As shown in Fig. \ref{fig:Interpretable}(a),
all heads are distributed with almost identical weights on different
tokens at the beginning of the training, reflecting the fact that
the agent doesn't have a clear concept of LTL instruction at the moment.
However, when the Transformer module converges, more weights focus on
the token $\mathsf{Black\_Zone}$ as shown in Fig. \ref{fig:Interpretable}(b),
which implies the agent having a greater probability of going directly
to the proposition $\mathsf{Black\_Zone}$. 

\section{CONCLUSIONS}

In this work, we present a T2TL framework that incorporates Transformer
to represent the LTL formula for improved performance and  interpretability. Future work will consider extensions
to multi-task learning.

\bibliographystyle{IEEEtran}
\bibliography{Bib4TL2}

\begin{thebibliography}{10}
\providecommand{\url}[1]{#1}
\csname url@samestyle\endcsname
\providecommand{\newblock}{\relax}
\providecommand{\bibinfo}[2]{#2}
\providecommand{\BIBentrySTDinterwordspacing}{\spaceskip=0pt\relax}
\providecommand{\BIBentryALTinterwordstretchfactor}{4}
\providecommand{\BIBentryALTinterwordspacing}{\spaceskip=\fontdimen2\font plus
\BIBentryALTinterwordstretchfactor\fontdimen3\font minus
  \fontdimen4\font\relax}
\providecommand{\BIBforeignlanguage}[2]{{%
\expandafter\ifx\csname l@#1\endcsname\relax
\typeout{** WARNING: IEEEtran.bst: No hyphenation pattern has been}%
\typeout{** loaded for the language `#1'. Using the pattern for}%
\typeout{** the default language instead.}%
\else
\language=\csname l@#1\endcsname
\fi
#2}}
\providecommand{\BIBdecl}{\relax}
\BIBdecl

\bibitem{Sutton2018}
R.~S. Sutton and A.~G. Barto, \emph{Reinforcement learning: An
  introduction}.\hskip 1em plus 0.5em minus 0.4em\relax MIT press, 2018.

\bibitem{vaswani2017attention}
A.~Vaswani, N.~Shazeer, N.~Parmar, J.~Uszkoreit, L.~Jones, A.~N. Gomez,
  {\L}.~Kaiser, and I.~Polosukhin, ``Attention is all you need,'' \emph{Adv.
  neural inf. process. syst}, vol.~30, 2017.

\bibitem{dosovitskiy2020image}
A.~Dosovitskiy, L.~Beyer, A.~Kolesnikov, D.~Weissenborn, X.~Zhai,
  T.~Unterthiner, M.~Dehghani, M.~Minderer, G.~Heigold, S.~Gelly \emph{et~al.},
  ``An image is worth 16x16 words: Transformers for image recognition at
  scale,'' \emph{arXiv preprint arXiv:2010.11929}, 2020.

\bibitem{chen2021decision}
L.~Chen, K.~Lu, A.~Rajeswaran, K.~Lee, A.~Grover, M.~Laskin, P.~Abbeel,
  A.~Srinivas, and I.~Mordatch, ``Decision transformer: Reinforcement learning
  via sequence modeling,'' \emph{Adv. neural inf. process. syst}, vol.~34, pp.
  15\,084--15\,097, 2021.

\bibitem{liu2022robot}
J.~Liu, Y.~Chen, Z.~Dong, S.~Wang, S.~Calinon, M.~Li, and F.~Chen, ``Robot
  cooking with stir-fry: Bimanual non-prehensile manipulation of semi-fluid
  objects,'' \emph{IEEE Robot. Autom. Lett.}, vol.~7, no.~2, pp. 5159--5166,
  2022.

\bibitem{Baier2008}
C.~Baier and J.-P. Katoen, \emph{Principles of model checking}.\hskip 1em plus
  0.5em minus 0.4em\relax MIT press, 2008.

\bibitem{Cai2021b}
M.~Cai, M.~Hasanbeig, S.~Xiao, A.~Abate, and Z.~Kan, ``Modular deep
  reinforcement learning for continuous motion planning with temporal logic,''
  \emph{IEEE Robot. Autom. Lett.}, vol.~6, no.~4, pp. 7973--7980, 2021.

\bibitem{Cai2021c}
M.~Cai, H.~Peng, Z.~Li, and Z.~Kan, ``Learning-based probabilistic ltl motion
  planning with environment and motion uncertainties,'' \emph{IEEE Trans.
  Autom. Control}, vol.~66, no.~5, pp. 2386--2392, 2021.

\bibitem{Li2019}
X.~Li, Z.~Serlin, G.~Yang, and C.~Belta, ``A formal methods approach to
  interpretable reinforcement learning for robotic planning,'' \emph{Sci.
  Robot.}, vol.~4, no.~37, 2019.

\bibitem{Icarte2022}
R.~T. Icarte, T.~Q. Klassen, R.~Valenzano, and S.~A. McIlraith, ``Reward
  machines: Exploiting reward function structure in reinforcement learning,''
  \emph{J. Artif. Intell. Res}, vol.~73, pp. 173--208, 2022.

\bibitem{Kuo2020}
Y.-L. Kuo, B.~Katz, and A.~Barbu, ``Encoding formulas as deep networks:
  Reinforcement learning for zero-shot execution of ltl formulas,'' in
  \emph{IEEE/RSJ Int. Conf. Intell. Robot. Syst.}, 2020, pp. 5604--5610.

\bibitem{vaezipoor2021ltl2action}
P.~Vaezipoor, A.~C. Li, R.~A.~T. Icarte, and S.~A. Mcilraith, ``Ltl2action:
  Generalizing ltl instructions for multi-task rl,'' in \emph{Int. Conf.
  Machin. Learn.}\hskip 1em plus 0.5em minus 0.4em\relax PMLR, 2021, pp.
  10\,497--10\,508.

\bibitem{yuan2022explainability}
H.~Yuan, H.~Yu, S.~Gui, and S.~Ji, ``Explainability in graph neural networks: A
  taxonomic survey,'' \emph{IEEE Trans. Pattern Anal. Mach. Intell.}, 2022.

\bibitem{Kupferman2001}
O.~Kupferman and M.~Y. Vardi, ``Model checking of safety properties,''
  \emph{Form. Methods Syst. Des.}, vol.~19, no.~3, pp. 291--314, 2001.

\bibitem{Mnih2015}
V.~Mnih, K.~Kavukcuoglu, D.~Silver, A.~A. Rusu, J.~Veness, M.~G. Bellemare,
  A.~Graves, M.~Riedmiller, A.~K. Fidjeland, G.~Ostrovski \emph{et~al.},
  ``Human-level control through deep reinforcement learning,'' \emph{Nature},
  vol. 518, no. 7540, pp. 529--533, 2015.

\bibitem{schulman2017}
J.~Schulman, F.~Wolski, P.~Dhariwal, A.~Radford, and O.~Klimov, ``Proximal
  policy optimization algorithms,'' \emph{arXiv preprint arXiv:1707.06347},
  2017.

\bibitem{ray2019benchmarking}
A.~Ray, J.~Achiam, and D.~Amodei, ``Benchmarking safe exploration in deep
  reinforcement learning,'' \emph{arXiv preprint arXiv:1910.01708}, vol.~7,
  p.~1, 2019.

\bibitem{ToroIcarte2018}
R.~Toro~Icarte, T.~Q. Klassen, R.~Valenzano, and S.~A. McIlraith, ``Teaching
  multiple tasks to an rl agent using ltl,'' in \emph{Proc. Int. Conf. Auton.
  Agents Multiagent Syst.}, 2018, pp. 452--461.

\bibitem{cai2023overcoming}
M.~Cai, E.~Aasi, C.~Belta, and C.-I. Vasile, ``Overcoming exploration: Deep
  reinforcement learning for continuous control in cluttered environments from
  temporal logic specifications,'' \emph{IEEE Robot. Autom. Lett.}, vol.~8,
  no.~4, pp. 2158--2165, apr 2023.

\bibitem{balakrishnan2022model}
A.~Balakrishnan, S.~Jaksic, E.~Aguilar, D.~Nickovic, and J.~Deshmukh,
  ``Model-free reinforcement learning for symbolic automata-encoded
  objectives,'' in \emph{HSCC - Proc. ACM Int. Conf. Hybrid Syst.: Comput.
  Control}, 2022, pp. 1--2.

\bibitem{kantaros2022}
Y.~Kantaros, ``Accelerated reinforcement learning for temporal logic control
  objectives,'' in \emph{IEEE/RSJ Int. Conf. Intell. Robot. Syst.}\hskip 1em
  plus 0.5em minus 0.4em\relax IEEE, 2022, pp. 5077--5082.

\bibitem{vig2020investigating}
J.~Vig, S.~Gehrmann, Y.~Belinkov, S.~Qian, D.~Nevo, Y.~Singer, and S.~Shieber,
  ``Investigating gender bias in language models using causal mediation
  analysis,'' \emph{Adv. neural inf. process. syst}, vol.~33, pp.
  12\,388--12\,401, 2020.

\bibitem{Vig19}
J.~Vig, ``A multiscale visualization of attention in the transformer model,''
  in \emph{ACL - Annu. Meet. Assoc. Comput. Linguist., Proc. Syst. Demonstr.},
  M.~R. Costa{-}juss{\`{a}} and E.~Alfonseca, Eds.\hskip 1em plus 0.5em minus
  0.4em\relax Association for Computational Linguistics, 2019, pp. 37--42.

\bibitem{zhang2021decision}
X.~Zhang, X.~Du, X.~Xie, L.~Ma, Y.~Liu, and M.~Sun, ``Decision-guided weighted
  automata extraction from recurrent neural networks.'' in \emph{AAAI}, 2021,
  pp. 11\,699--11\,707.

\bibitem{araki2021learning}
B.~Araki, K.~Vodrahalli, T.~Leech, C.-I. Vasile, M.~Donahue, and D.~Rus,
  ``Learning and planning with logical automata,'' \emph{AUTON ROBOT}, vol.~45,
  no.~7, pp. 1013--1028, 2021.

\bibitem{li2021learning}
X.~Li, G.~Rosman, I.~Gilitschenski, B.~Araki, C.-I. Vasile, S.~Karaman, and
  D.~Rus, ``Learning an explainable trajectory generator using the automaton
  generative network (agn),'' \emph{IEEE Robot. Autom. Lett.}, vol.~7, no.~2,
  pp. 984--991, 2021.

\bibitem{Wang2016}
J.~X. Wang, Z.~Kurth-Nelson, D.~Tirumala, H.~Soyer, J.~Z. Leibo, R.~Munos,
  C.~Blundell, D.~Kumaran, and M.~Botvinick, ``Learning to reinforcement
  learn,'' \emph{arXiv preprint arXiv:1611.05763}, 2016.

\bibitem{Rakelly2019}
K.~Rakelly, A.~Zhou, C.~Finn, S.~Levine, and D.~Quillen, ``Efficient off-policy
  meta-reinforcement learning via probabilistic context variables,'' in
  \emph{Int. Conf. Mach. Learn.}\hskip 1em plus 0.5em minus 0.4em\relax PMLR,
  2019, pp. 5331--5340.

\bibitem{FakoorCSS20}
R.~Fakoor, P.~Chaudhari, S.~Soatto, and A.~J. Smola, ``Meta-q-learning,'' in
  \emph{Int. Conf. Learn. Represent.}, 2020.

\bibitem{Lacerda2014}
B.~Lacerda, D.~Parker, and N.~Hawes, ``Optimal and dynamic planning for markov
  decision processes with co-safe ltl specifications,'' in \emph{IEEE/RSJ Int.
  Conf. Intell. Robot. Syst.}\hskip 1em plus 0.5em minus 0.4em\relax IEEE,
  2014, pp. 1511--1516.

\bibitem{VanHasselt2016}
H.~Van~Hasselt, A.~Guez, and D.~Silver, ``Deep reinforcement learning with
  double q-learning,'' in \emph{Proc. AAAI Conf. Artif. Intell.}, vol.~30,
  no.~1, 2016.

\end{thebibliography}

\end{document}